\documentclass{article} 
\usepackage{iclr2021_conference,times}

\usepackage{amsmath,amsfonts,bm}

\def\eqref#1{equation~\ref{#1}}

\def\1{\bm{1}}

\DeclareMathAlphabet{\mathsfit}{\encodingdefault}{\sfdefault}{m}{sl}
\SetMathAlphabet{\mathsfit}{bold}{\encodingdefault}{\sfdefault}{bx}{n}

\usepackage{graphicx}

\usepackage{hyperref}
\usepackage{url}

\title{Imperfect ImaGANation: Implications of GANs Exacerbating Biases on Facial Data}

\author{
$\text{Niharika Jain*}^{1}, \text{Alberto Olmo*}^{1}, \text{Sailik Sengupta}^{1}, \text{Lydia Manikonda}^{2}, \text{Subbarao Kambhampati}^{1}$\\
$^{1}$Arizona State University \ \ $^{2}$Rensselaer Polytechnic Institute \\
$^*$\texttt{\{njain30, aolmoher\}@asu.edu} \\
}

\iclrfinalcopy 
\begin{document}

\maketitle

\begin{abstract}
In this paper, we show that popular Generative Adversarial Networks (GANs) exacerbate biases along the axes of gender and skin tone when given a skewed distribution of face-shots. While practitioners celebrate synthetic data generation using GANs as an economical way to augment data for training data-hungry machine learning models, it is unclear whether they recognize the perils of such techniques when applied to real world datasets biased along latent dimensions. Specifically, we show that (1) traditional GANs further skew the distribution of a dataset consisting of engineering faculty headshots, generating minority modes less often and of worse quality and (2) image-to-image translation (conditional) GANs also exacerbate biases by lightening skin color of non-white faces and transforming female facial features to be masculine when generating faces of engineering professors. Thus, our study is meant to serve as a cautionary tale.

\end{abstract}

\section{Introduction}

The use of Generative Adversarial Networks (GANs) \citep{goodfellow2014generative} has grown significantly and due to data-demand of deep learning models, when faced with sparse data (owing to paywalls, privacy concerns, etc.) practitioners often turn to promising data augmentation solutions. While earlier computer vision works focused on performing affine transformations to existing samples \citep{o1995document,bloice2017augmentor}, using GANs for synthetic data generation has recently become popular \citep{forbes-synthetic,techcrunch-synthetic}. GANs generate such data by approximating the original distribution with a limited training set and create examples that appear novel. These examples give a (false) sense of sampling unseen data from the same underlying distribution as the original training data, making GANs a seemingly perfect candidate for data augmentation. We note that even this best-case scenario would be a territory for practitioners to tread lightly; GAN-generated data for augmentation would only propagate the existing biases of the real-world data. Owing to theoretical limitations of GANs \citep{arora2017gans}, we show a grim reality: the generated data learns a distribution shifted from that of the real world, one which exacerbates these biases and disproportionately underrepresents those already in the minority, both in number and quality. This poses serious ethical implications on any downstream tasks trained on a synthetically-augmented dataset, especially when biases exist along protected or embargoed attributes.

\section{Architecture and Approach}


\noindent \textbf{Mode Collapse~~}
GANs are known to estimate an equilibrium of a minimax game played by a generator network $G$ and discriminator network $D$. While $D$, a binary classifier, learns to discriminate between images that come from a real-world data distribution $p_{data}$ and those that do not, $G$ learns to generate images from $p_{GAN}$ and fool $D$ into classifying them as coming from $p_{data}$.
In the presence of infinite training data, computation time and network capacity for the generator and the discriminator, this process ensures that the $p_{GAN}$ distribution generated by $G$ converges to that of the training data $p_{data}$ \citep{goodfellow2014generative}. In reality, GAN-generated distributions are not nearly as diverse as their training distributions \citep{arora2017gans,arora2017generalization} and the support (i.e. possible feature combinations of the generated data) is only representative of a small subset of what one would expect to see when sampling data from the real distribution. The support size of the generated images is constrained by the capacity of $D$.
$G$ collapses because the set of noise inputs that would correspond to some minority mode in the image space has (by definition) a low probability of being seen by $D$. As $G$ only optimizes its own weights over the feedback from $D$, it rarely learns to generate these modes \citep{che2016mode}.

\looseness=-1
There are several related works-- \cite{zhao2018bias} studies GANs' bias and generalization to unseen modes without discussing the problem of GANs collapsing to existing modes. While mode collapse is a well-studied phenomenon \citep{grnarova2018evaluating,goodfellow2016nips,che2016mode,arora2017generalization}
and several GAN variants have been developed to alleviate its effects \citep{metz2016unrolled,srivastava2017veegan,arjovsky2017wasserstein,miyato2018spectral,tolstikhin2017adagan,karras2017progressive},
a distinction is rarely made between uniform and non-uniform training datasets. On these lines, \cite{mishra2018mode} empirically shows that the divergence between $p_{data}$ and $p_{GAN}$ does indeed worsen as the training data is more skewed, however using four scalar metrics which do not offer much insight on \emph{how} the distributions differ. 
For a dataset that is biased along latent axes (e.g. gender and skin color), we hypothesize $G$ (we try several GAN variants) collapses to modes in the majority groups (e.g. masculine and white faces) amplifying biases that exist in the original data.

\noindent \textbf{Data Collection and Processing~~}
To test our hypothesis, we construct a dataset of faces of engineering professors from U.S. universities that are (1) listed in the top 47 of US News' most recent ``Best Engineering Schools'' and (2) had public access to faculty directories with images. The data exhibits bias along the latent dimensions of gender and race and thus, is an appropriate test-bed to study the amplification of bias in GAN-based data generation.
We gather a total of $17,245$ engineering faculty $64\times64$-pixel headshots using an unsupervised face detector \citep{Dalal05}.

\section{Experiment and Results}


\begin{figure}[!t]
\centering
\includegraphics[width=\textwidth]{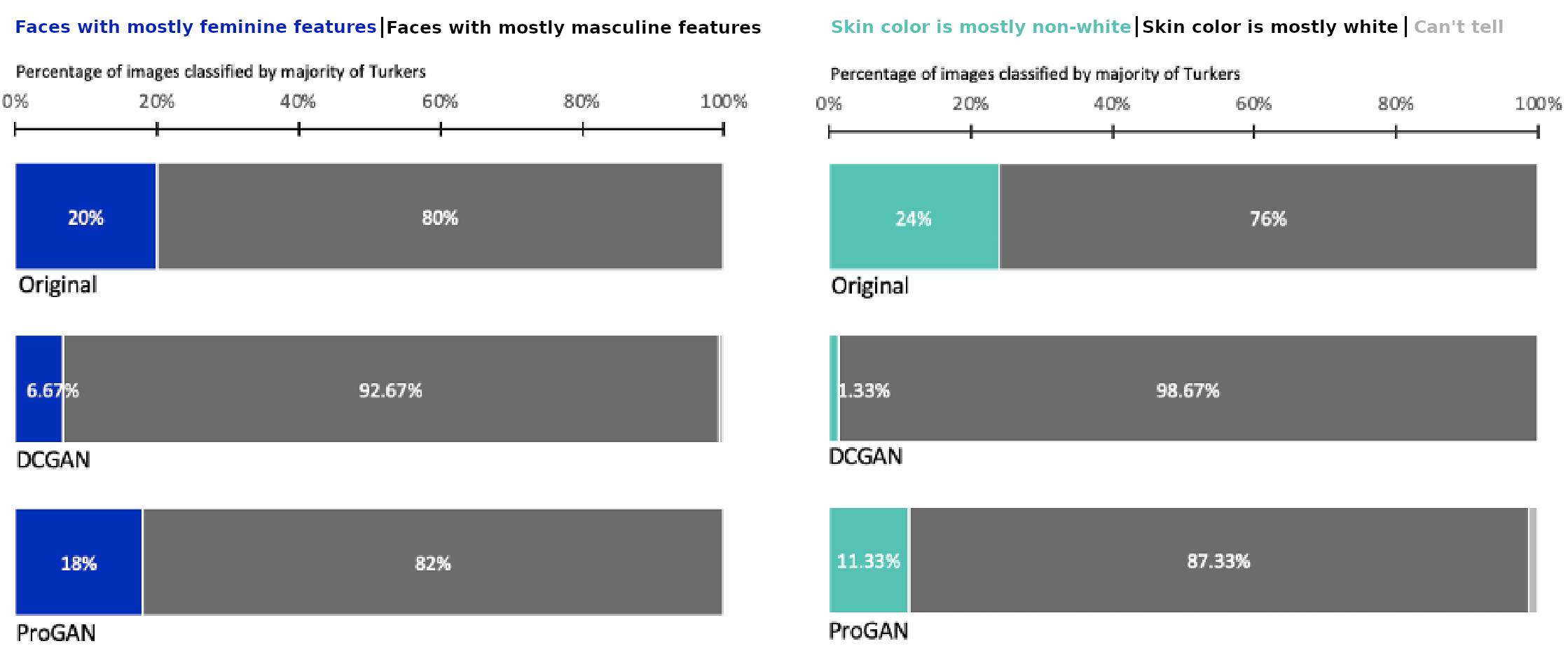}
\caption{Distribution of human classifications on gender and skin color.}
\label{fig:turkerStudies}
\end{figure}

To explore the diversity of $p_{GAN}$ we test the performance on three GANs (1) DCGAN \citep{radford2015unsupervised}: the most common GAN used by practitioners due to its minimal requirements for compute power and off-the-shelf availability \citep{dcgan-repo}, (2) ProGAN \citep{karras2017progressive}: a state-of-the-art GAN for sample quality and known to addresses the mode-collapse problem and to overcome the quality-variance tradeoff \citep{karras2019style,karras2020analyzing}, and (3) CycleGAN \citep{zhu2017unpaired}: the most well-known image-to-image translation GAN, which transforms an image from one domain to another by minimizing cycle-consistency and identity losses. We show experiments on two other GAN architectures designed to address mode collapse -- Wasserstein GAN \citep{arjovsky2017wasserstein} and AdaGAN \citep{tolstikhin2017adagan} -- in the appendix.


\vspace{-0.15cm}
\subsection{Imagining Engineers from Scratch}
We assess the data from the GAN variants -- DCGAN and ProGAN -- by asking humans to annotate images from the original and generated datasets along the dimensions of race and gender. To account for variance in model training, we generate $50$ images from three seeds where each seed trains the DCGAN and the ProGAN for $50$ epochs. 
We conduct 4 seven-minute human study tasks in a between-subject design fashion (each annotator saw images belonging to only one set) and leveraged data from $234$ master Turkers on Amazon's MTurk platform. Each worker performed the tasks:

\noindent [\texttt{T1(a/b)}] Human subjects were asked to select the most appropriate option for an image $x$ sampled from [{\tt T1a}] $p_{data}$ and [{\tt T1b}] $G(z)$ with the following options: (1) face mostly has masculine features, (2) face mostly has feminine features, and (3) neither of the above is true.

\noindent [\texttt{T2(a/b)}] Human subjects were asked to select the most appropriate option for an image $x$ sampled from [{\tt T2a}] $p_{data}$ and [{\tt T2b}] $G(z)$ from the list of following options: (1) skin color is non-white, (2) skin color is white, and (3) can't tell.

We presented each annotator with $52$ images-- $50$ from the original/generated data and two high quality trivial images with known labels for gender and skin color. This helped us prune $18$ bad datapoints. We had $30$ valid data points for all generated datasets and $25$ for the original distribution. We considered majority-voting to categorize an image as belonging to a class. \autoref{fig:turkerDisag} in the appendix contains the resulting charts.

\subsubsection{Results}
We plot the results for {\tt T1a} and {\tt T1b} in \autoref{fig:turkerStudies} (left) and find that (1) both DCGAN and ProGAN penalize the original $20\%$ of images with mostly feminine features being DCGAN the most penalizing, reducing the percentage to $6.67\%$. A one-tailed two-proportion z-test yields a p-value of $0.0032$ confirming the amplification of bias across the latent dimension of gender for DCGAN and (2) for tasks {\tt T2a} and {\tt T2b} (\autoref{fig:turkerStudies}, right)
the proportion of non-white faces decreased from $24\%$ in the original dataset to $1.33\%$ in the DCGAN-generated dataset and to $11.33\%$ for ProGAN. The p-value obtained ($2.7\times10^{-8}$ for DCGAN and $1.05\times10^{-3}$ for ProGAN) show strong statistical significance as both GANs collapse along the latent dimension of race, biasing the synthetic faces toward lighter skin tones. Note that while ProGAN did not collapse along the axis of gender, it was not immune to collapsing along the axis of other protected features (eg. skin color). We notice that the synthetic data not only propagates but exacerbates those biases against minority populations.

\noindent \textbf{Quality and Confidence Metrics}
We measure the consensus among Turkers by the amount of votes needed to classify each image in the axes of gender and color. For DCGAN, we find that the proportion of images labelled as non-white and female decreases as the voting threshold increases. This is indicative of a higher level of agreement between participants and shows that the quality of generated images for the minority classes is worse than that of the majority classes. ProGAN does not exhibit this disparity in quality across gender, but it produces lower quality for non-white faces than white ones.

\begin{figure}[t]
\centering
\includegraphics[width=\columnwidth]{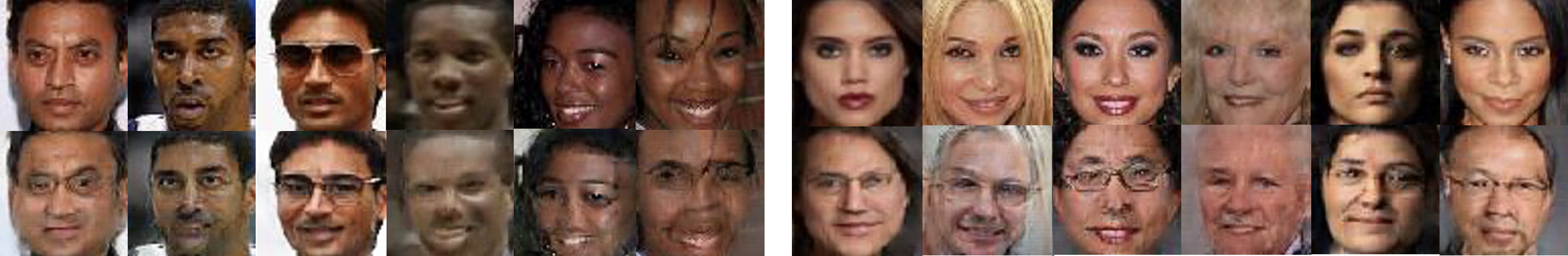}
\caption{Illustrative test set of transformations on non-white (two rows on the left) and female celebrities (two rows on the right). Original and stylized images are one atop another respectively. }
\label{fig:cyclegan}
\end{figure}

\subsection{Imagining Engineering Counterparts}
As image-to-image translation GANs' output distributions conditioned on the input, our intuition was that they may be less susceptible to exacerbating biases. For instance, in our task where gender is a latent feature and feminine faces are underrepresented, a GAN, provided with the input image of a female, would have to actively convert it into a male one. Unfortunately, it is known that even these conditional GAN variants are not immune to mode collapse \citep{ma2018gan}. However, how conditional variants of GANs react to sensitive social features such as race and gender remains an open question.

To study this, we train a CycleGAN \citep{zhu2017unpaired} to stylize faces of non-engineering professors to look like engineering faculty. Thus, our target/output domain consists of the engineering faculty face dataset leveraged in the previous experiment and our input domain is the CelebA dataset \citep{liu2015faceattributes} consisting of over 200,000 annotated images of celebrities. As our dataset consists of only 16,500 images, we randomly sample 16,500 faces from the CelebA dataset for training. We then create a held-out test set from CelebA in which we have 100 images for each of the four categories-- white, non-white, male, and female.

In \autoref{fig:cyclegan}, we showcase the transformation of celebrity faces that are representative of the minority categories (i.e. non-white, female) in the engineering professors dataset. While we see that the GAN learns to add glasses or creating smiling expressions, not all the modifications learned are socially harmless, we also notice that it lightens the skin tone of non-white celebrities and imparts masculine aspects to the faces of female celebrities.
While it is reasonable to expect that a GAN might perpetuate and exacerbate biases along any arbitrary dimension where there exists a skew in the training set, we stress that this kind of innocuous bias is not our focus. Machine learning systems are designed to find correlations to recognize patterns, but this correlation-seeking becomes problematic for social features when models perpetuate and exacerbate biases for minority groups who have faced systemic disadvantage or discrimination. Before concluding, we highlight a case-study where such models are having adverse real-world impact.

\vspace{-0.3cm}
\section{Real-World Applications and Conclusion}
While our experiments meant to serve as example, the bias-exacerbation consequences of mode collapse in GANs can be seen in real-world applications. Snapchat, a popular image-sharing platform, has recently taken 
advantage of the image-to-image translation capabilities of conditional GANs such as CycleGAN for their "My Twin" lens, according to several sources \citep{yanjiali_2020,magazine_2019,jang_1970,reddit}. We show that this presumably conditional-GAN-based technology reacts to the sensitive features this work discusses. When applying this lens to a female face, the GAN should ideally make no changes, but when used on women of color, it lightens skin tone, though this is not the case for white women using the same filter. While we have not performed a comprehensive study, the observations and claims open an intriguing research problem \citep{baeza2016data}. Examples of the lightened complexions on women of color and white women can be seen in \autoref{fig:fake_real_faces} in the appendix.

The implications of using a biased facial dataset augmented via GANs for a downstream task could be severe. The use of machine learning models on facial data is already prevalent in critical decision-making scenarios such as employment \citep{hymas_2019}, healthcare \citep{bahrampour_2014}, education \citep{kaur_marco_2019}, criminal justice \citep{harwell_2019}, as well as security innovations like deepfake detection \citep{DFDC}. It is of clear ethical import that we ensure our training sets and models are fair and diverse with respect to sensitive features. At the very least, they ought not to rig the system \emph{against} already underrepresented minorities.

GANs have proven to create less diverse distributions than the original they are trained on, but 
the implications of mode collapse remain unclear in scenarios where the training distribution $p_{data}$ is biased toward certain feature values (eg. males) along a latent feature (eg. gender). To study this, we empirically show how GANs trained on a demographic already skewed toward white and male faces exacerbate social biases in the generated distribution $p_{GAN}$. In our setting, mode collapse occurs on a majority latent mode of the original data and causes a severe under-representation of feminine facial features and non-white skin tones in the generated dataset. We also demonstrate that this perpetuation of biases against female and non-white features occurs in image-to-image translation GANs, first stylizing celebrities' faces to look like those of engineering professors, and next by conducting a case study on Snapchat's "My Twin" lens.
Beyond implications about social issues, this work should serve as a general caution against using GAN-based data augmentation techniques to alleviate problems arising from sparse or unbalanced datasets for any downstream task.
There seems to exist a false sense of security that GANs can generate {\em novel} data samples which pick the expected semantic features relating to the defect, and place them in previously unseen settings. In actuality, the augmented data might be underrepresenting or compromising image quality for some crucial feature of the real-world data.

\bibliography{iclr2021_conference}
\bibliographystyle{iclr2021_conference}

\newpage
\appendix
\section{Appendix}

\begin{figure*}
    \centering
    \includegraphics*[width=0.9\linewidth]{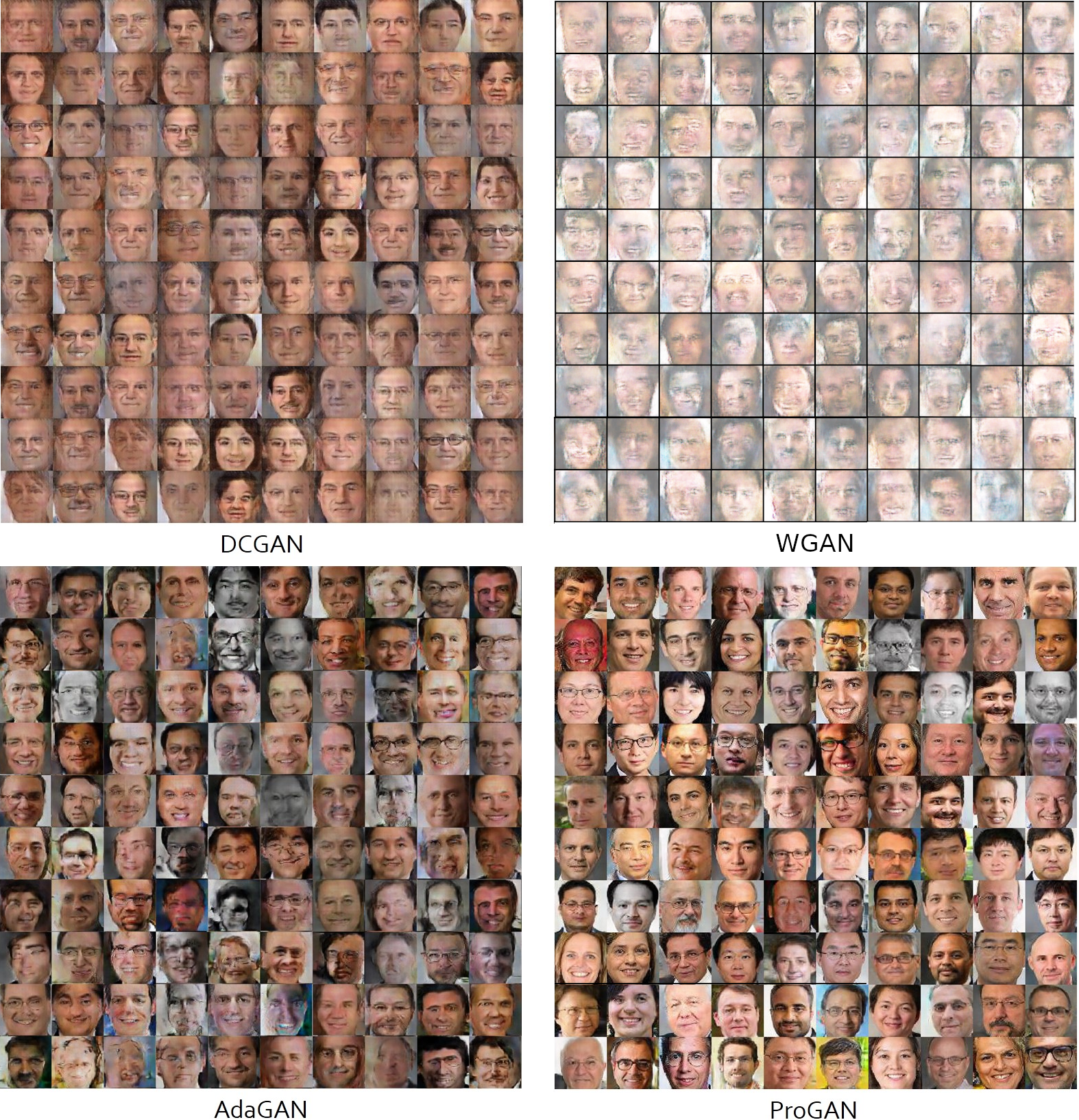}
    \caption{Images of professors generated by popular GAN architectures trained on our engineering professors dataset; WGAN, AdaGAN, and ProGAN attempt to address the mode-collapse problem.}
    \label{fig:scenario1}
\end{figure*}

In addition to the variants mentioned in the paper (DCGAN and ProGAN), we investigate the performance of another two GANs which claim to reduce mode collapse: \textit{Wasserstein} GANs or WGAN \citep{arjovsky2017wasserstein} and AdaGAN \cite{tolstikhin2017adagan}.

\begin{figure}[!t]
\centering
\includegraphics[width=0.8\textwidth]{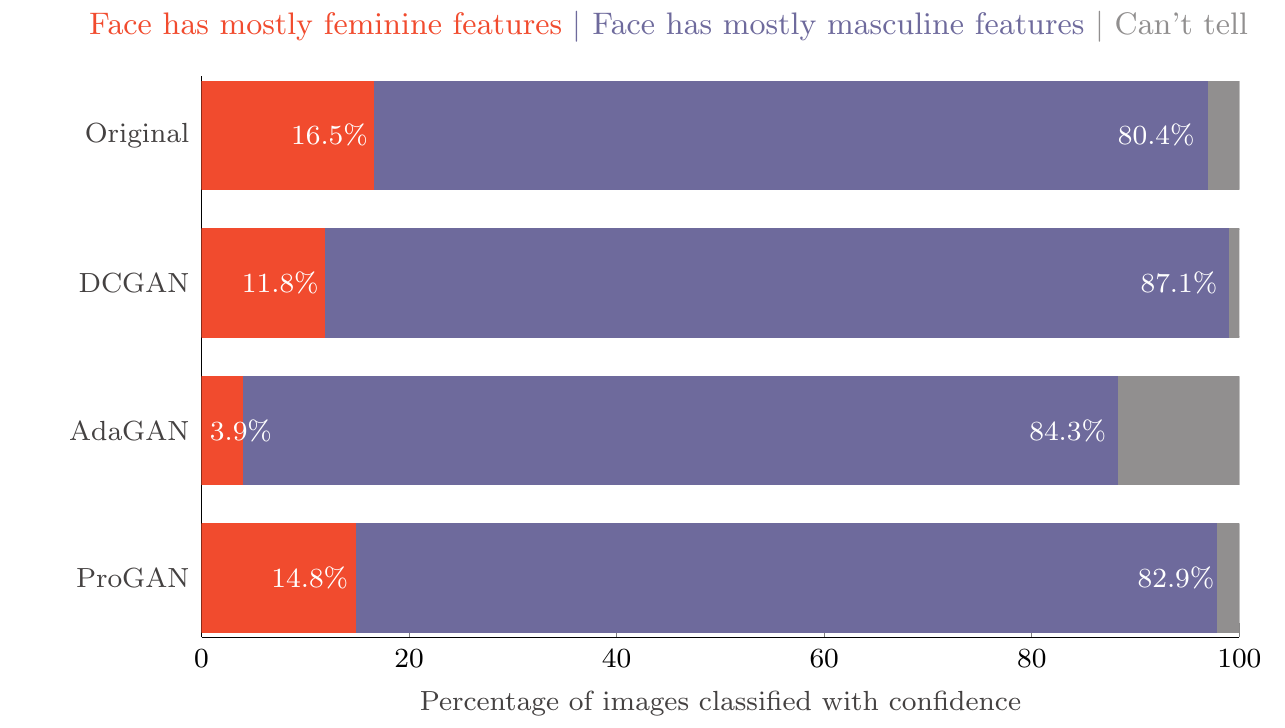}
\caption{The percentage of faces classified as female, male and can't tell by Microsoft Azure's Face API decreases from $16.5\%$ in the original dataset significantly in the synthetically generated datasets across several GAN variants that are popular or attempt to address the mode collapse problem.}
\label{fig:microsoftGender}
\end{figure}

\subsection{Studies with Commercially Available Gender Classification Systems}

As a less subjective approach to labelling, we use Microsoft Azure's Face API for classifying $5000$ images from the training set and $5000$ generated by the three different variants of GAN: the popular DCGAN and two others that attempt to address the problem of mode collapse, AdaGAN and ProGAN (we omit WGAN due to poor image quality). To ensure our results are not specific to a single generation, we obtain $5000$ images by sampling from three runs of each GAN with different random seeds for weight initialization. We show 100 images randomly sampled from these $5000$ obtained from each GAN variant in \autoref{fig:scenario1}.

In \autoref{fig:microsoftGender}, we show the percentage of images classified as female, male and ``can't'' tell by Microsoft's AI tool. We perform a one-tailed two-proportion z-test on the original and generated distributions to assess the null hypothesis that the proportion of feminine features in the synthetic distribution for all GAN variants, is same as in the original. In the initial dataset, $16.5\%$ are labelled as females while $80.4\%$ are labelled as male, clearly indicating an original bias towards males. Further, DCGAN exacerbates it significantly (with a p-value of $3.2\times
10^{-6}$), bringing down the percentage of females in the generated set to $11.8\%$. The quality of images generated by AdaGAN is significantly worse than the ones produced by all other variants as indicated by the spike in the number of images, from $3.1\%$ in the original to $11.8\%$ (with a p-value of $3.0\times10^{-53}$). Surprisingly, regardless of the poor quality, there is a significant increase in the number of generated images that are classified as male (from $80.4\%$ in the original data to $84.3\%$) while the number of generated images that are classified as females has a substantial drop (from $16.5\%$ to $3.9\%$). This also highlights that many of the other GAN variants that seek to address mode-collapse, have proven to be worse than AdaGAN \citep{lala2018evaluation}; such as WGAN \citep{arjovsky2017wasserstein}, VEEGAN \citep{srivastava2017veegan} or Unrolled GAN \citep{metz2016unrolled} and either affect the quality of generated images, exacerbate the biases over latent features such as gender, or both. On the other hand, the more recent architecture ProGAN, clearly outperforms both the popular DCGAN and AdaGAN by reducing the exacerbation of bias and improving image quality. It only decreases the percentage of females in its generated set by $1.7\%$, even though this is a significant exacerbation of bias along the latent dimension of gender (with a p-value of $0.09008$). Our results show that popular and state-of-the-art GAN variants paint an optimistic picture of this technology for data-augmentation while suffering from the exacerbation of biases along latent dimensions.

\begin{figure}[!t]
\centering
\includegraphics[width=\textwidth]{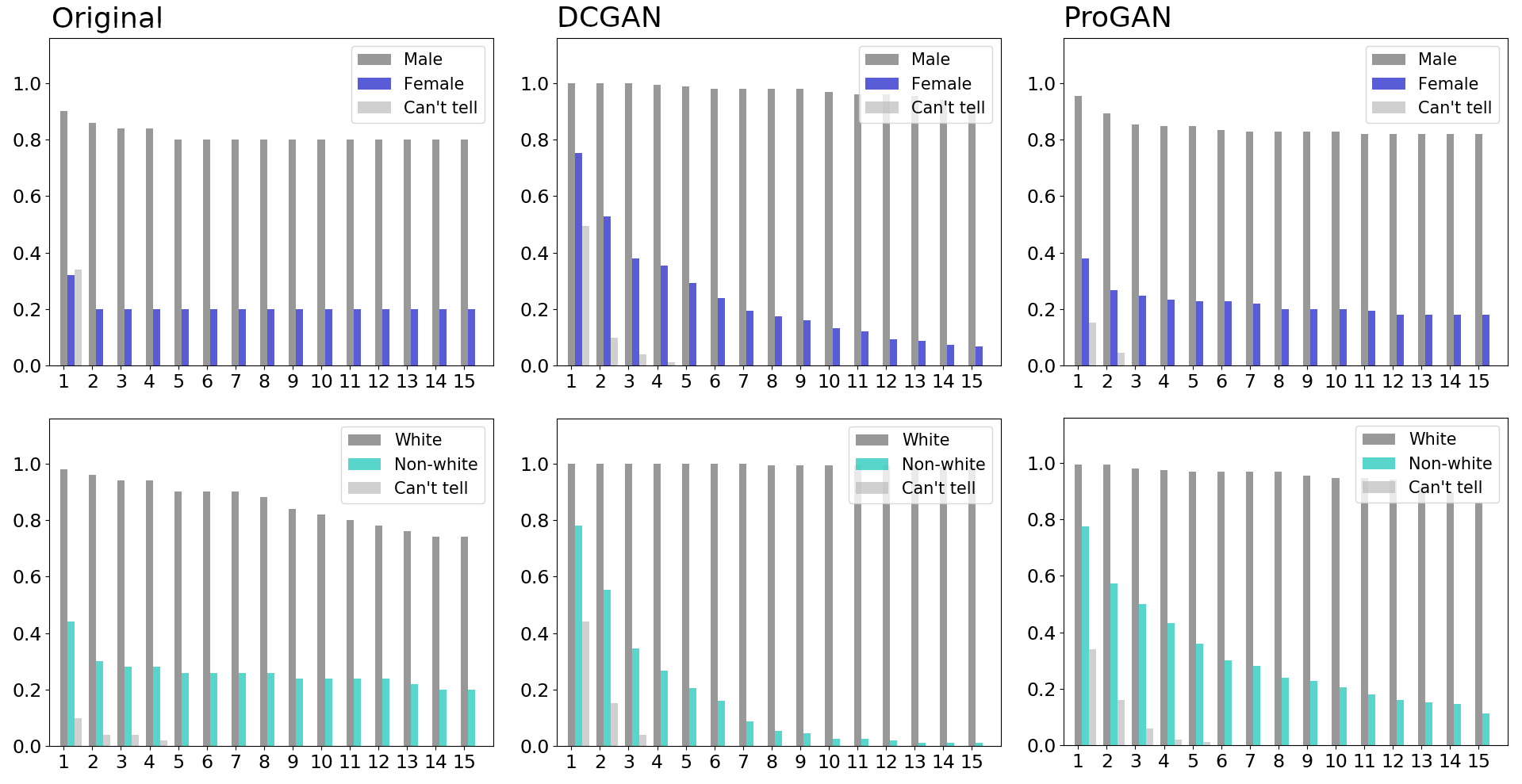}
\caption{Human annotator agreements on skin color and gender between professor headshots from the original and synthetic (generated by DCGAN and ProGAN) distributions. The number of images labeled as masculine, feminine or neither, changes as the threshold number of votes required to categorize an image into a particular category increases from $1$ to $15$.}
\label{fig:turkerDisag}
\end{figure}

\subsection{Confidence Metrics for DCGAN and ProGAN}
We measure the confidence of the annotators for the synthetic datasets by plotting how each headshot is classified when the threshold varies from $1$ to $15$ and show it in \autoref{fig:turkerDisag}. The x-axis represents the number of votes required to classify an image to each particular class (i.e. male, female, ``can't tell''), and the y-axis represents the proportion of images that are classified considering that voting threshold.
This is a metric for consensus among the group of annotators that the images belong to a certain class and demonstrates their annotating confidence; the less variability through y over each threshold, the more confidence the workers show. On the original data, roughly the same proportion of images are classified as male, female, white, and nonwhite irrespective of the number of Turkers needed to vote. In other words, the Turkers are confident about which faces are male, female, white, and non-white. This is not the case for the synthetic distributions. For DCGAN, the proportion of images that are marked as female or non-white significantly drops as it requires more Turkers to vote for that label, and they only lose confidence over the images depicting the minority gender and race; the proportion of images marked as male or white does not drop as the voting threshold increases. For ProGAN's images, the Turkers are confident about male, female, and white faces, but not about non-white faces.

\subsection{Snapchat Case Study}
Image-to-image translation GANs, such as pix2pix or CycleGAN \citep{isola2017image,zhu2017unpaired} adjust colors and textures in an already-existing image from some input domain to map it to another class. Normally, the input and target domains are closely related and the mapping can be achieved by changing the geometries minimally. Some examples of successful applications for image-to-image translation are conversion of horses to zebras, street photographs to their semantic segmentation, aerial photos to Google maps, and summer landscapes to winter landscapes. CycleGAN is the most popular off-the-shelf GAN variant used by machine learning practitioners today, as measured by the number of stars on the most-used GitHub repositories for this model \citep{cyclegan-repo,cyclegan-repo-2}, and has also, predictably, been a popular choice for synthetic data augmentation \citep{301c889c2fb74be89d5afc1e3efbd4fb,sandfort2019data,huang2018auggan}.
Just as with the unconditional variants, our motivation is to explore if and how the diversity of the generated distribution $p_{GAN}$ differs from the training distribution $p_{data}$.

\begin{figure}[h]
\centering
\includegraphics[width=\columnwidth]{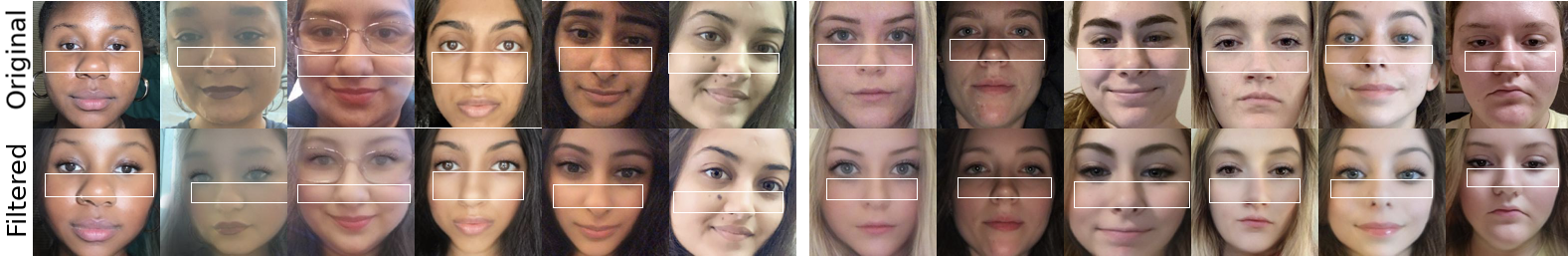}
\caption{Faces of women of color (left six columns) and white women (right six columns) before and after using Snapchat's female gender face lens, top and bottom respectively. The sections used for the skin-color machine analysis are highlighted in white.}
\label{fig:fake_real_faces}
\end{figure}

To assess how the skin color changed between pairs of images objectively, we crop a section of the face under the eyes and above the tip of the nose, spanning both cheeks, 
and find its average pixel value, then we map the RGB vector, using L2-norm distance, to the closest standard shade in the L'Or\'eal skin color chart\footnote{\url{https://www.loreal.com/en/articles/science-and-technology/expert-inskin/}}.
While not considering skin warmth, only skin lightness, we show that the
lens lightens non-white faces by one shade consistently for five faces and produces no effect for one of them. On the other hand, it performed randomly for white faces in our example, lightening two by one shade, darkening two by one shade, and not affecting two. 
A potential cause of lightening skin tones in women of color is that a GAN used by the face lens collapses all inputs in a region of the image space to output lighter colors. However, more rigorous studies should be performed to make certain claims. Our case study offers initial support for the narrative of Snapchat's beautification face lenses lightening skin tones for people of color. 





\subsection{Downstream Tasks and Vulnerable Communities}
The glaring ethical problem with automated, machine learning-powered tools is that they are ``most often used on people towards whom they exhibit the most bias,'' and that the errors arising from bias ``can be much more costly for those in marginalized communities than other groups'' \citep{gebru2019oxford}. Classification tasks in the real world suffer from this dilemma. In criminal justice, automated tools predict recidivism risk in a system which disproportionately punishes Black and Hispanic people. It is unfortunate yet unsurprising, then, that the risk assessment software used in state criminal justice systems is biased against Black people \citep{propublica}. This classification system is input over 137 features -- not including race -- and disproportionately classifies Black defendants as medium or high risk. In employment, automated tools predict candidate performance and fit in industries which are already male-dominated. Further, a hiring system designed by Amazon in 2018 faced public backlash when it was found to discriminate against female candidates by penalizing r\'esum\'es which included participation in women's organizations. The classification system scraped r\'esum\'es of candidates from the past ten years and was never given gender as an input feature. Classifiers are not the only automated tool who would use data generated by GANs. In 2020, PULSE \citep{menon2020pulse}, a face ``depixelizer,'' received widespread backlash on social media (especially from world-famous contributors to the field of AI ethics) because it was shown to upsample images of non-white faces to have Caucasian features. The authors of this paper perform several studies in response and conclude that the biases in PULSE derive directly from the biased performance of the GAN from which it receives generated data. The major takeaway of all discussions mentioned here is that the data bias problem cannot be reduced solely to the dataset used. It seems that popular automated data generation tools, namely GANs, will not merely perpetuate the patterns found in the data (the theoretical ideal for the technologies), but rather amplify them. The question is, then, how we can regulate the societal applications for which these known flawed systems are used.

\end{document}